\documentclass{kapproc}

\usepackage{graphicx}
\usepackage{dcolumn}
\usepackage{bm}
  \usepackage{procps} 
\normallatexbib

\newcommand{\highlite}[1]{\smallskip\noindent{\bf\underline{#1}}}

\begin{document}

\articletitle{Extremal Optimization: an Evolutionary Local-Search Algorithm}

\author{Stefan Boettcher\footnote{www.physics.emory.edu/faculty/boettcher/}}
\affil{Department of Physics, Emory University, Atlanta, GA 30322, USA}
\email{sboettc@emory.edu}

\author{Allon G. Percus\footnote{www.c3.lanl.gov/\~{}percus/}}
\affil{Computer and Computational Sciences Division, MS-B265, Los Alamos National Laboratory, Los Alamos, NM 87545, USA}
\email{percus@lanl.gov}

\begin{abstract}
A recently introduced general-purpose heuristic for finding
high-quality solutions for many hard optimization problems is
reviewed. The method is inspired by recent progress in understanding
far-from-equilibrium phenomena in terms of {\em self-organized
criticality,\/} a concept introduced to describe emergent complexity
in physical systems.  This method, called {\em extremal
optimization,\/} successively replaces the value of extremely
undesirable variables in a sub-optimal solution with new, random
ones. Large, avalanche-like fluctuations in the cost function
self-organize from this dynamics, effectively scaling barriers to
explore local optima in distant neighborhoods of the configuration
space while eliminating the need to tune parameters. Drawing upon
models used to simulate the dynamics of granular media, evolution, or
geology, extremal optimization complements approximation methods
inspired by equilibrium statistical physics, such as {\em simulated
annealing}. It may be but one example of applying new insights into
{\em non-equilibrium phenomena} systematically to hard optimization
problems.  This method is widely applicable and so far has proved
competitive with -- and even superior to -- more elaborate
general-purpose heuristics on testbeds of constrained optimization
problems with up to $10^5$ variables, such as bipartitioning,
coloring, and satisfiability. 
 Analysis of a
suitable model predicts the only free parameter of the method in
accordance with all experimental results.
\end{abstract}
\begin{keywords}
Combinatorial Optimization, Heuristic Methods, Evolutionary Algorithms, Self-Organized Criticality.
\end{keywords}

\section{Introduction}
\label{background}
Extremal optimization (EO) \cite{BoPe1,GECCO,CISE} is a
general-purpose local search heuristic based on recent progress in
understanding far-from-equilibrium phenomena in terms of
self-organized criticality (SOC) \cite{BTW}. It was inspired by
previous attempts of using physical intuition to optimize, such as
simulated annealing (SA) \cite{Science} or genetic algorithms
\cite{Goldberg}. It opens the door to systematically applying {\em
non-equilibrium processes\/} in the same manner as SA applies
equilibrium statistical mechanics. EO appears to be a powerful
addition to the above mentioned Meta-heuristics \cite{Osman} in its
generality and its ability to explore complicated configuration spaces
efficiently.

Despite original aspirations, even conceptually elegant methods such
as SA or GA did not provide a panacea to optimization. The incredible
diversity of problems, few resembling physics, just would not allow
for that. Hence, the need for creative alternatives arises. We will
show that EO provides a {\em true\/} alternative approach, with its
own advantages and disadvantages, compared to other general-purpose
heuristics. It may not be the method of choice for many problems; a
fate shared by all methods. Based on the existing studies, we believe
that EO will prove as indispensable for some problems as other
general-purpose heuristics have become.

In the next section, we will motivate EO in terms of the evolutionary
model by Bak and Sneppen \cite{BS1}. In Sec.~\ref{EOalgorithm}, we
discuss the general EO-implementation on the example of graph
bipartitioning. Finally, in Sec.~\ref{EOapplications}, we describe
implementations for other problems and some results we have obtained.

\section{Bak-Sneppen Model}
\label{BSmodel}
The EO heuristic was motivated by the Bak-Sneppen model of biological
evolution \cite{BS1}. In this model, ``species'' are located on the
sites of a lattice (or graph), and have an associated ``fitness''
value between 0 and 1.  At each time step, the one species with the
smallest value (poorest degree of adaptation) is selected for a random
update, having its fitness replaced by a new value drawn randomly from
a flat distribution on the interval $[0,1]$.  But the change in
fitness of one species impacts the fitness of interrelated species.
Therefore, all of the species connected to the ``weakest'' have their
fitness replaced with new random numbers as well. After a sufficient
number of steps, the system reaches a highly correlated state known as
self-organized criticality (SOC) \cite{BTW}.  In that state, almost
all species have reached a fitness above a certain threshold.  These
species possess {\em punctuated equilibrium\/} \cite{G+E}: only one's
weakened neighbor can undermine one's own fitness.  This
coevolutionary activity gives rise to chain reactions or
``avalanches'': large (non-equilibrium) fluctuations that
rearrange major parts of the system, potentially making any
configuration accessible.

Although coevolution does not have optimization as its exclusive goal,
it serves as a powerful paradigm for EO \cite{BoPe1}. EO follows the
spirit of the Bak-Sneppen model in that it merely updates those
variables having an extremal (worst) arrangement in the current
configuration, replacing them by random values without ever explicitly
improving them. Large fluctuations allow to escape from local minima to
efficiently explore the configuration space, while the extremal
selection process enforces frequent returns to near-optimal
configurations. This selection {\em against} the ``bad'' contrasts
sharply with the ``breeding'' pursued in GAs.

\section{Extremal Optimization Algorithm}
\label{EOalgorithm}
Many practical decision-making problems can be modeled and analyzed in
terms of standard combinatorial optimization problems, the most
intractable ones provided by the class of NP-hard problems
\cite{G+J}. These problems are considered {\em hard\/} to solve
because they require a computational time that in general grows faster
than any power of the number of variables, $n$, in an instance to
discern the optimal solution, in close analogy to many real-world
optimization problems \cite{Rayward}.  Study of such problems has
spawned the development of {\em efficient\/} \cite{Cook} approximation
methods called {\em heuristics,} i.~e. methods that find approximate,
near-optimal solutions rapidly \cite{Reeves}.

\begin{figure}
\vskip 1.7truein \includegraphics{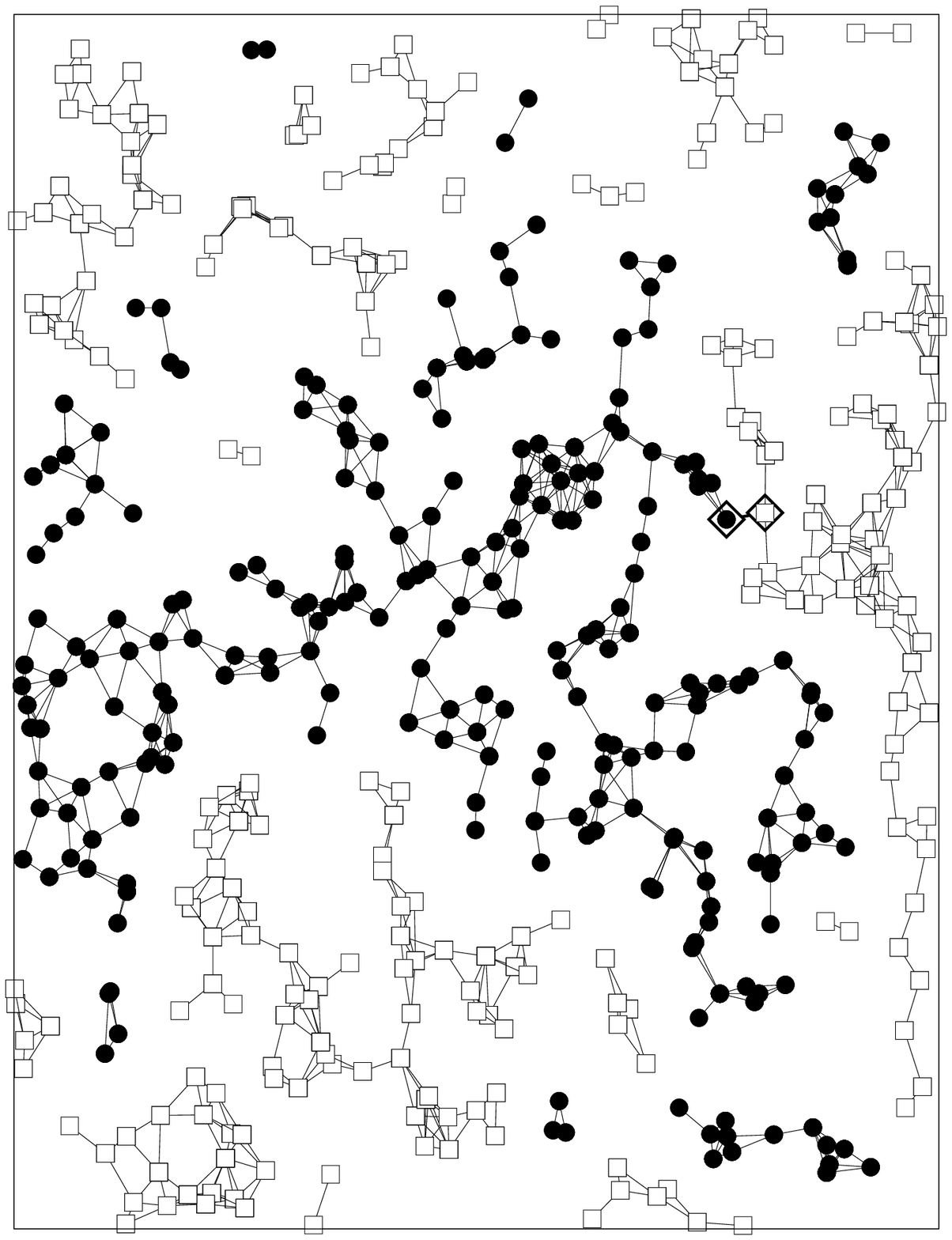} \includegraphics{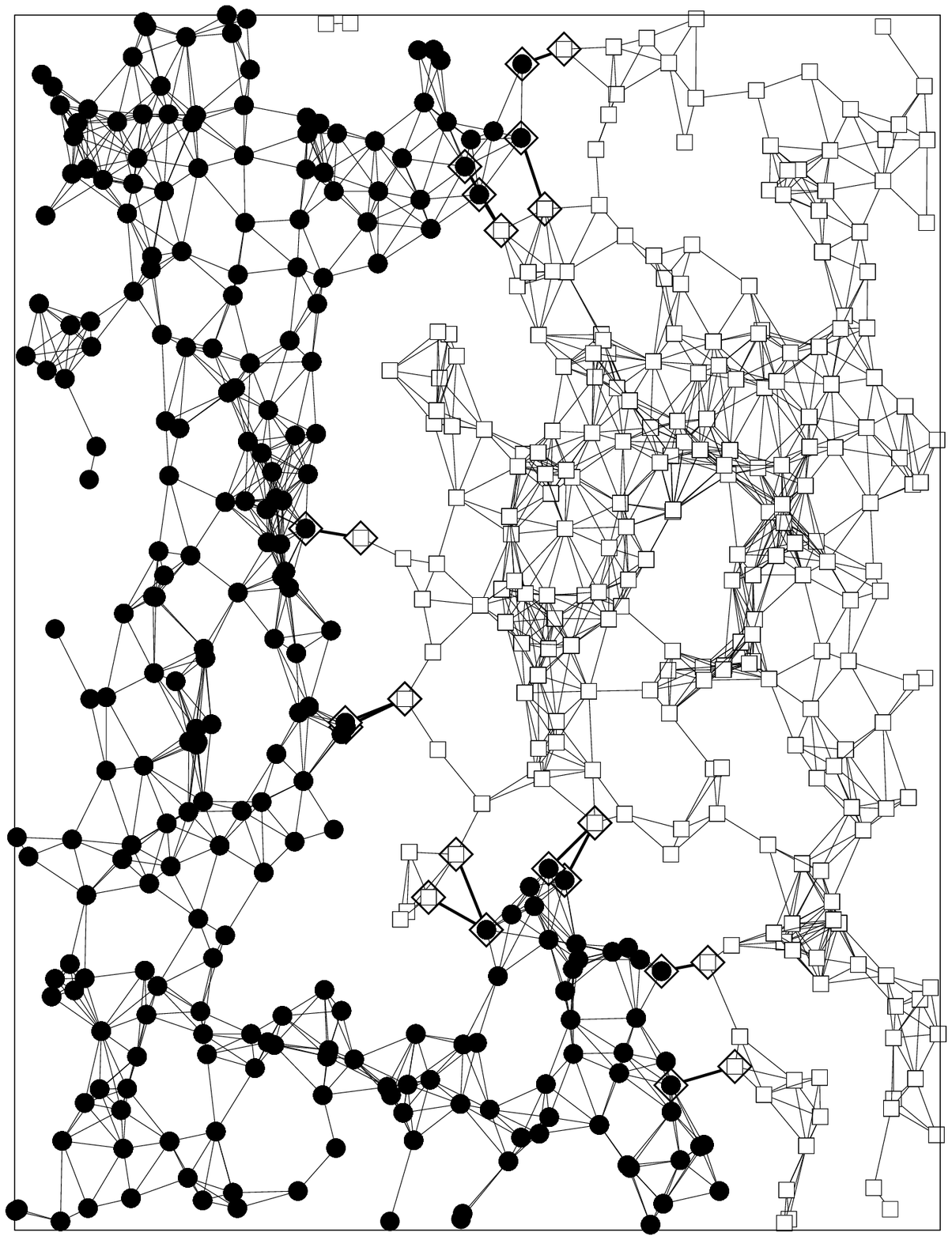}
\caption{Two random geometric graphs, $n=500$, with connectivity 4
(top) and connectivity 8 (bottom) in an optimized configuration found
by EO. At $\alpha=4$ the graph barely ``percolates,'' with merely one
``bad'' edge (between points of opposite sets, masked by diamonds)
connecting a set of 250 round points with a set of 250 square
points. For the denser graph on the bottom, EO reduced the cutsize to
13. A 1-$exchange$ will turn a square vertex into a round one, and
a round vertex into a square one.  }
\label{geograph}
\end{figure}

One example of a hard problem with constraints is the graph
bi-partitioning problem (GBP)~\cite{G+J,Science,JohnsonGBP},
see Fig~\ref{geograph}.
Variables $x_i$ are given by a set of $n$ vertices, where $n$ is
even. ``Edges'' connect certain pairs of vertices to form an instance
of a graph. The problem is to find a way of partitioning the vertices
into two subsets, each constrained to be {\em exactly\/} of size
$n/2$, with a {\em minimal\/} number of edges between the subsets. In
the GBP, the size of the configuration space $\Omega$ grows
exponentially with $n$, $|\Omega|=\left(n\atop n/2\right)$, since all
unordered divisions of the $n$ vertices into two equal-sized sets are
feasible configurations $S\in\Omega$. The cost function $C(S)$
(``cutsize'') counts the number of ``bad'' edges that need to be cut
to separate the subsets.  A typical local-search neighborhood $N(S)$
for the GBP arises from a ``1-$exchange$'' of one vertex from each
subset, the simplest update that preserves the global constraint.

To find near-optimal solutions on a hard problem such as the GBP, EO
performs a search on a single configuration $S\in\Omega$ for a
particular optimization problem. Characteristically, $S$ consists of a
large number $n$ of variables $x_i$. Theses variables usually can
obtain a state from a set $I$ which could be Boolean (as for the GBP
or $K$-SAT), $p$-state (as for $p$-partitioning or $p$-coloring), or
continuous (similar to the Bak-Sneppen model above). We assume that
each $S$ possesses a neighborhood $N(S)$, originating from updates of
some of the variables. The cost $C(S)$ is assumed to be a linear
function of the ``fitness'' $\lambda_i$ assigned to each variable
$x_i$ (although that is not essential \cite{BoPe1}). Typically, the
fitness $\lambda_i$ of variable $x_i$ depends on its state in relation
to other variables that $x_i$ is connected to. Ideally, it is
\begin{eqnarray}
C(S)=-\sum_{i=1}^n \lambda_i.
\label{costeq}
\end{eqnarray}
For example, in the GBP, Eq.~(\ref{costeq}) is satisfied, if we
attribute to each vertex $x_i$ a local cost $\lambda_i=-b_i/2$, where
$b_i$ is the number of its ``bad'' edges, equally shared with the
vertex on the other end of that edge. On each update, a vertex $x_j$
is identified which possesses the lowest fitness $\lambda_j$. (If more
than one vertex has lowest fitness, the tie is broken at random.) A
neighboring configuration $S'\in N(S)$ is chosen
via the 1-$exchange$ 
by swapping $x_j$ with a randomly selected vertex from the opposite
set.

For minimization problems, EO proceeds as follows:
\smallskip
\begin{center}
\framebox[3.8in]{
\begin{minipage}[t]{3.5in}
\begin{enumerate}
\item Initialize configuration $S$ at will; set $S_{\rm best}\!:=\!S$.
\item For the ``current'' configuration $S$,
\label{EOupdate}
\begin{enumerate}
\item evaluate $\lambda_i$ for each variable $x_i$,
\label{evaluate}
\item find $j$ satisfying $\lambda_j\leq\lambda_i$ for all $i$, {\em
i.e.,\/} $x_j$ has the ``worst fitness'',
\label{sort}
\item choose $S'\!\in\!N(S)$ such that $x_j$ {\em must} change,
\label{worst}
\item accept $S:=S'$ {\em unconditionally,}
\label{alwaysmove}
\item if $C(S)<C(S_{\rm best})$ then set $S_{\rm best}:=S$.
\end{enumerate}
\item Repeat at step~(\ref{EOupdate}) as long as desired.
\item Return $S_{\rm best}$ and $C(S_{\rm best})$.
\end{enumerate}
\end{minipage}}
\end{center}
\smallskip
The algorithm operates on a single configuration $S$ at each
step. Each variable $x_i$ in $S$ has a fitness, of which the ``worst''
is identified. This {\em ranking\/} of the variables provides the only
measure of quality on $S$, implying that all other variables are
``better'' in the current $S$. In the move to a neighboring
configuration $S'$, typically only a small number of variables change
state, such that only a few connected variables need to be
re-evaluated [step~(\ref{evaluate})] and re-ranked
[step~(\ref{sort})].  Note that there is not a single parameter to
adjust for the selection of better solutions aside from this
ranking. In fact, it is the {\em memory\/} encapsulated in this
ranking that directs EO into the neighborhood of increasingly better
solutions.  On the other hand, in the choice of move to $S'$, there is
no consideration given to the outcome of such a move, and not even the
worst variable $x_j$ itself is guaranteed to improve its
fitness. Accordingly, large fluctuations in the cost can accumulate in
a sequence of updates. Merely the bias {\em against\/}
extremely``bad'' fitnesses enforces repeated returns to near-optimal
solutions.

\begin{figure}
\vskip 2.2truein \includegraphics{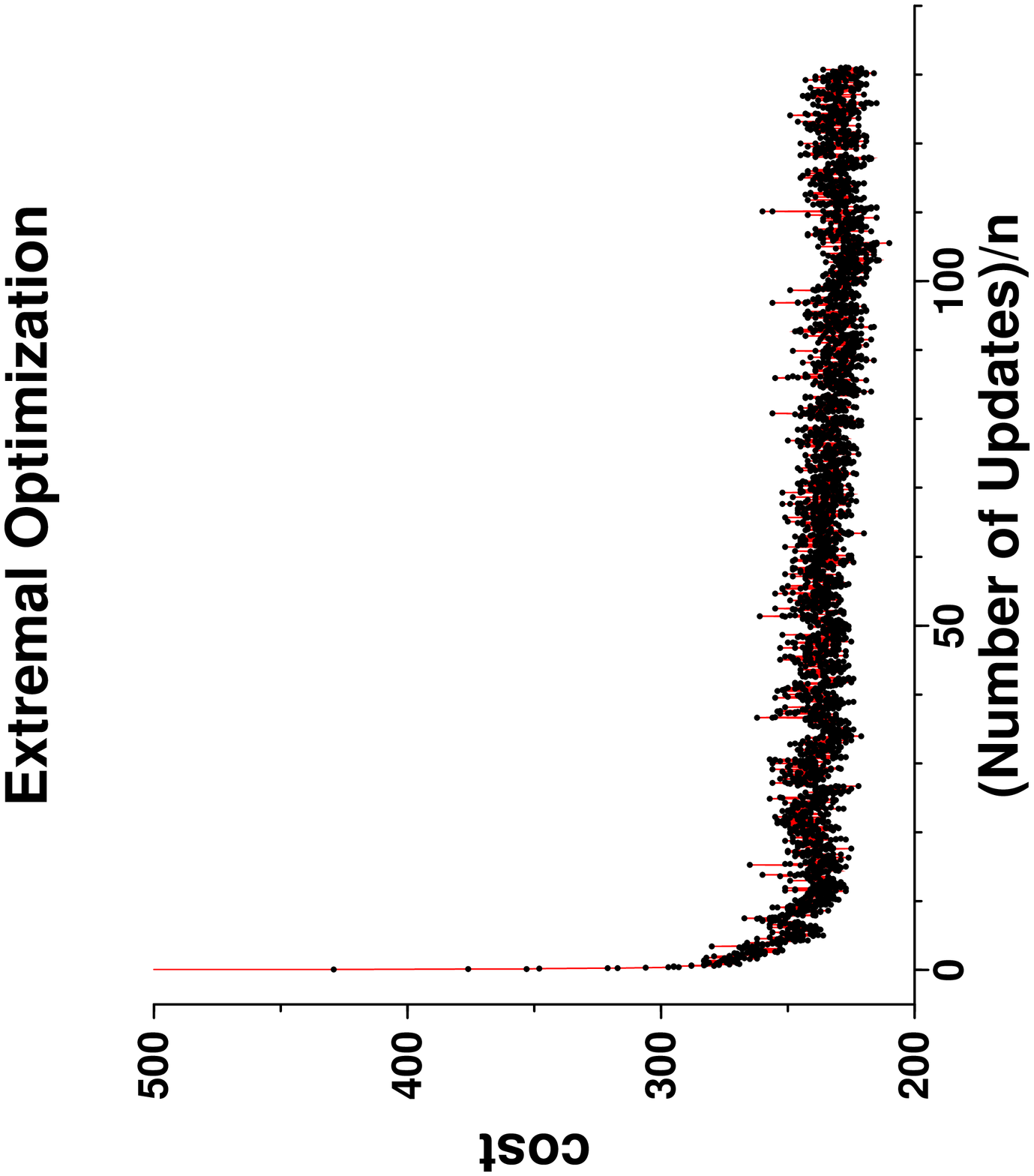} \includegraphics{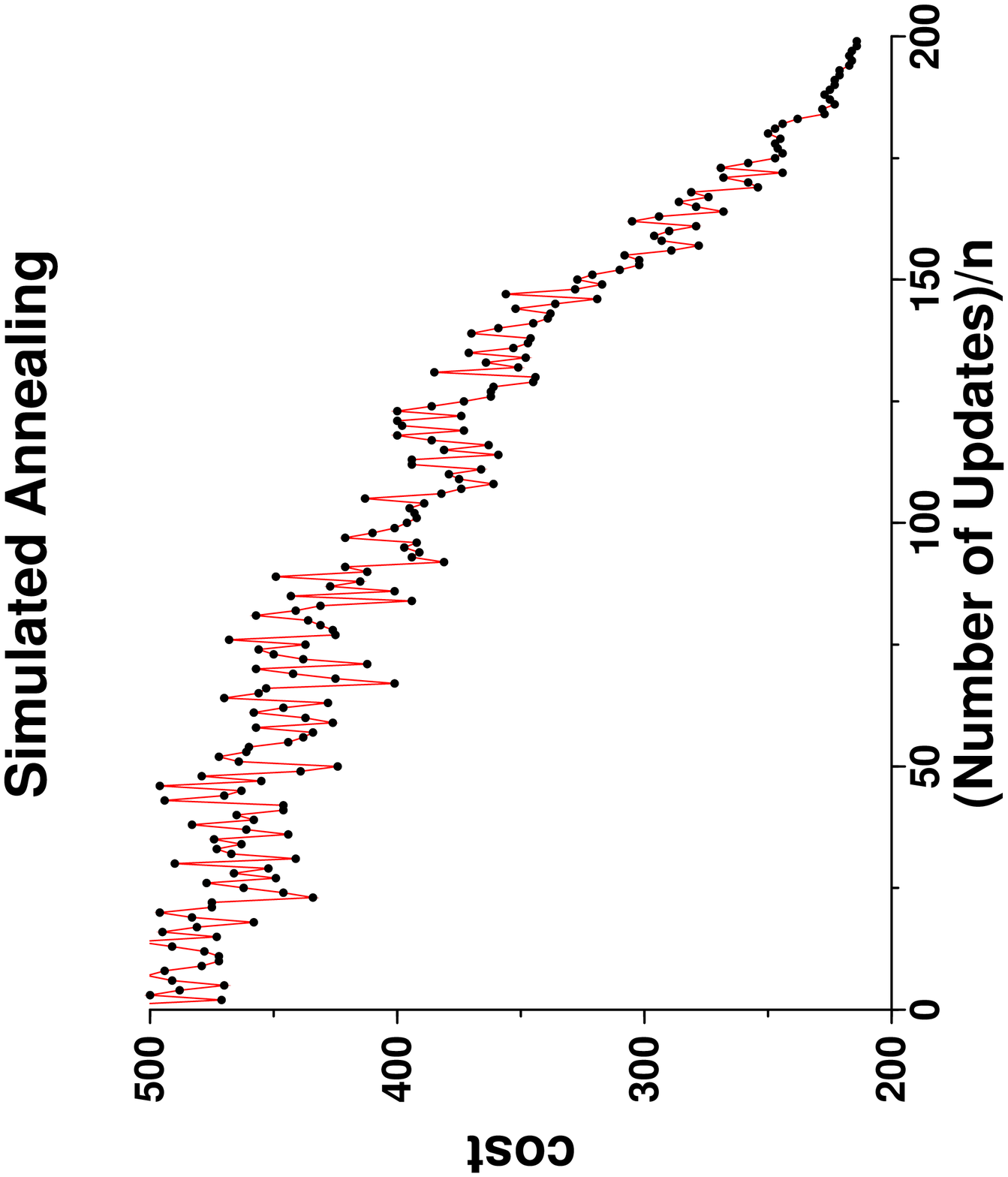}
\caption{Evolution of the cost function $C(S)$ during a typical run of
 EO (left) and  SA (right) for the bipartitioning of an $n=500$-vertex graph
$G_{500}$ introduced in Ref.~\protect\cite{JohnsonGBP}. The best cost
ever found for $G_{500}$ is 206.  In contrast to SA, which has large
fluctuations in early stages of the run and then converges much later,
extremal optimization quickly approaches a stage where broadly
distributed fluctuations allow it to probe and escape many local
minima.  }
\label{runtime}
\end{figure}

A typical ``run'' of this algorithm for the GBP \cite{BoPe1} is
shown in Fig.~\ref{runtime}. It illustrates that near-optimal
configurations are often revisited, although large fluctuations abound
even in latter parts of the run.


\subsection{$\tau$-EO Algorithm}
\label{taueoalgo}
Tests have shown that this basic algorithm is very competitive for
optimization problems where EO can choose randomly among many $S'\in
N(S)$ that satisfy step~(\ref{worst}) such as for the GBP \cite{BoPe1}.  But, as we will see below, sometimes the neighborhood
$N$ chosen for a problem turns EO into a deterministic process:
selecting always the worst variable in step~(\ref{sort}) leaves no
choice in step~(\ref{worst}). Like {\em iterative improvement}, such
an EO-process would get stuck in local minima.  To avoid these ``dead
ends,'' and to improve results generally\cite{BoPe1}, we introduce a
single parameter into the algorithm. This parameter, $\tau$, remains
fixed during each run and varies for each problem
only with the system size $n$.

The parameter $\tau$ allows us to exploit the memory contained in the
fitness ranking for the $x_i$ in more detail.  We find a permutation
$\Pi$ of the labels $i$ with
\begin{eqnarray}
\lambda_{\Pi(1)}\leq\lambda_{\Pi(2)}\leq\ldots\leq\lambda_{\Pi(n)}.
\label{rankeq}
\end{eqnarray}
The worst variable $x_j$ [step~(\ref{sort})] is of rank 1, $j=\Pi(1)$,
and the best variable is of rank $n$. Now, consider a probability
distribution over the {\em ranks\/} $k$,
\begin{eqnarray}
P_k\propto k^{-\tau},\qquad 1\leq k\leq n,
\label{taueq}
\end{eqnarray}
for a given value of the parameter $\tau$. At each update, select a
rank $k$ according to $P_k$. (For sufficiently large $\tau$, this procedure will again select the vertex with the worst fitness, $k=1$, but for any finite $\tau$, it will occasionally dislodge fitter variables, $k>1$.) Then, modify step~(\ref{sort}) so that
the variable $x_j$ with $j=\Pi(k)$ gets chosen for an update in
step~(\ref{worst}). For example, in the case of the GBP with a
1-$exchange$, we now select {\em two\/} numbers $k_1$ and $k_2$
according to $P_k$ and swap vertex $j_1=\Pi(k_1)$ with vertex
$j_2=\Pi(k_2)$ (we repeat drawing $k_2$ until $j_1$ and $j_2$ are from
opposite sets).

\begin{figure}
\vskip 2.0truein \includegraphics{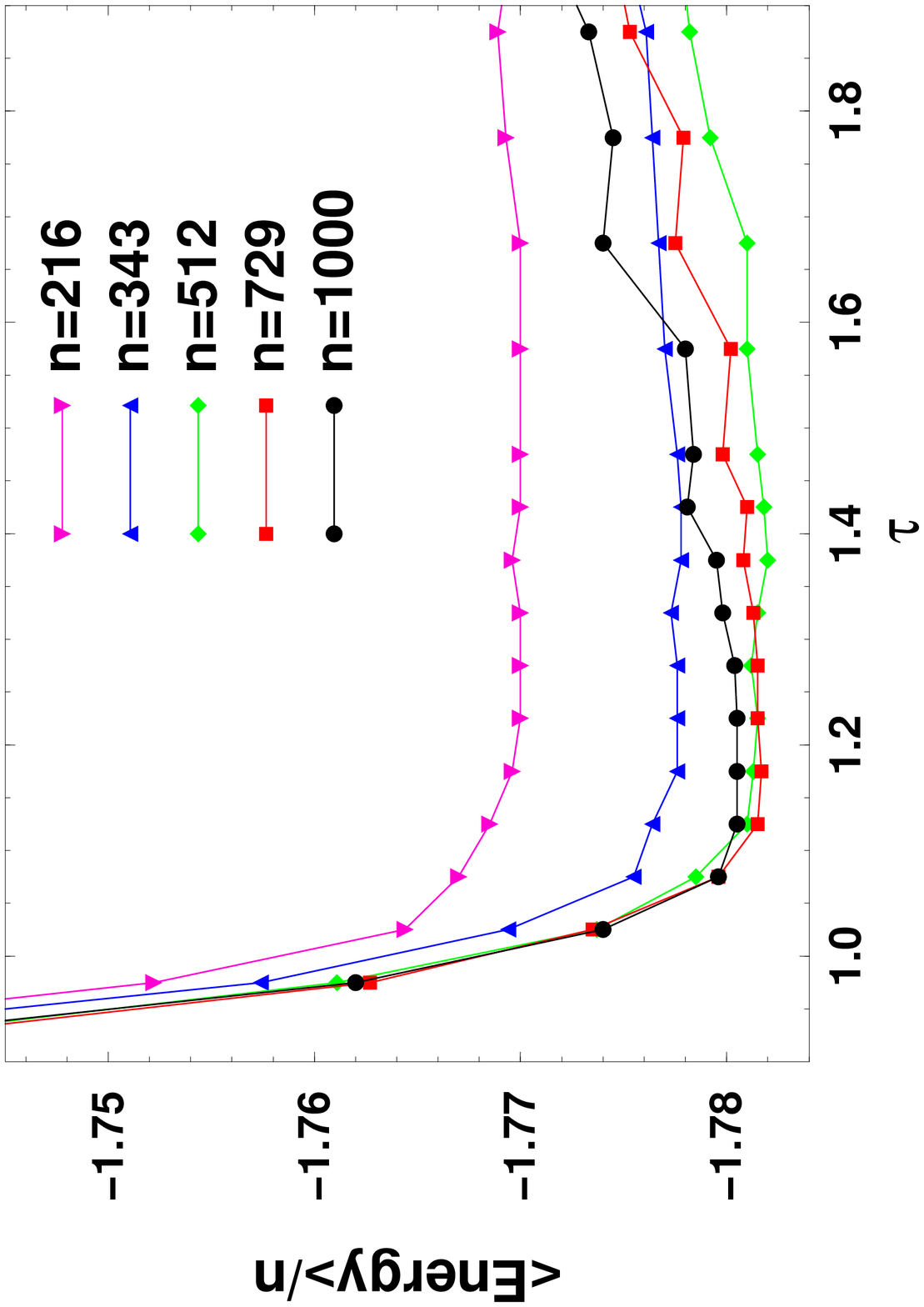} \includegraphics{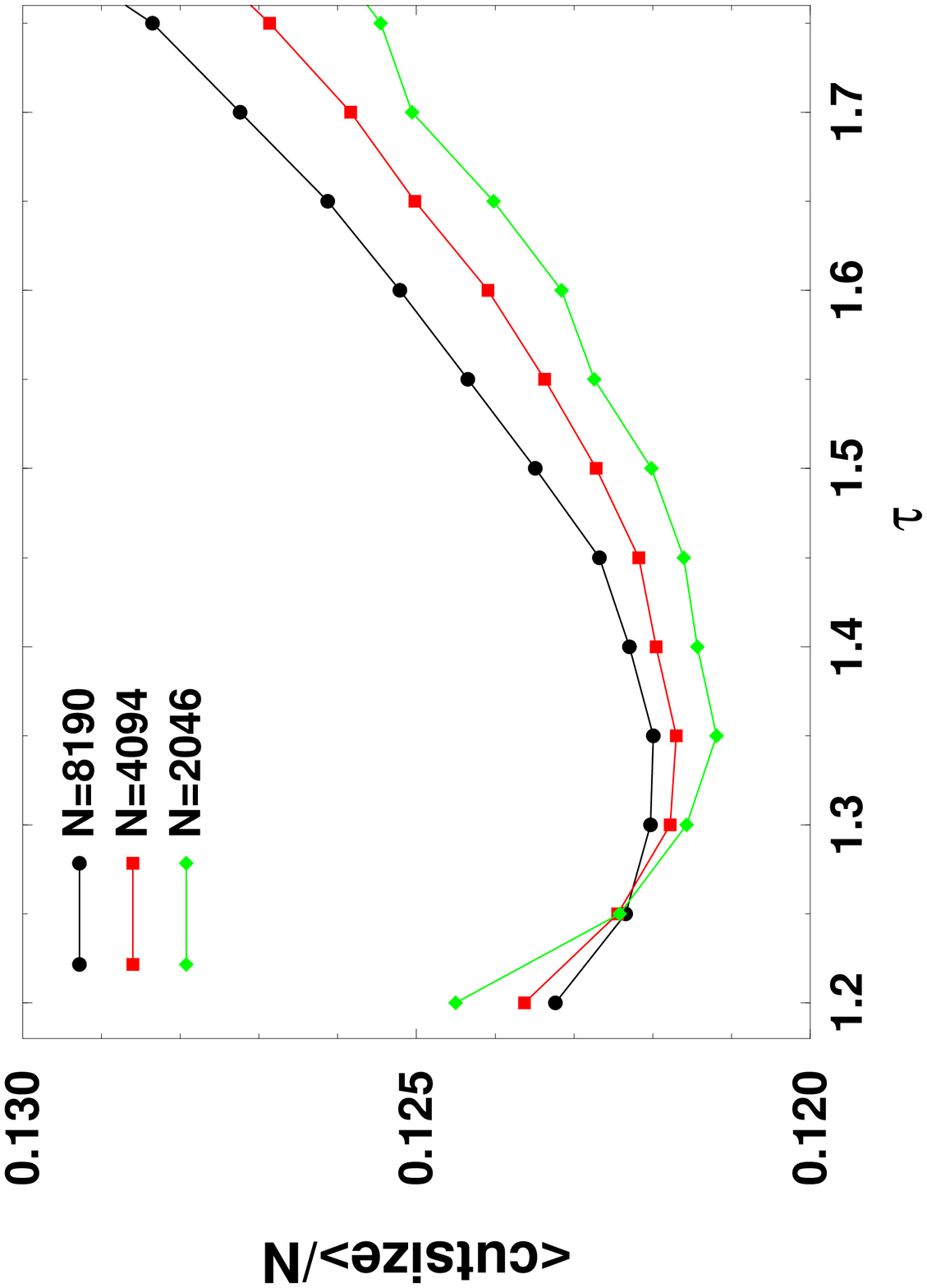}
\caption{Plot of the average costs obtained by EO  for a $\pm J$
spin glass (left) and for graph bipartitioning (right), as a
function of $\tau$.  For each size $n$, a number of instances were
generated. For each instance, 10 different EO runs were performed at
each $\tau$.  The results were averaged over runs and over instances.
Although both problems are quite distinct, in either case the best
results are obtained for
$\tau\to1^+$ for $n\to\infty$. }
\label{glasstau}
\end{figure}

For $\tau=0$, this ``$\tau$-EO'' algorithm is simply a local random
walk through $\Omega$.  Conversely, for $\tau\to\infty$, the process
can approach a deterministic local search, only updating the
lowest-ranked variable(s), and may be bound to reach a dead end (see
Fig.~\ref{glasstau}). In both extremes the results are typically
poor. However, for intermediate values of $\tau$ the choice of a
(scale-free) power-law distribution for $P_k$ in Eq.~(\ref{taueq})
ensures that no rank gets excluded from further evolution, while still
maintaining a bias against variables with bad fitness. As we will show in the next section, the
$\tau$-EO algorithm can be analyzed to show that an asymptotic choice
of $\tau-1\sim[\ln(n)]^{-1}$ optimizes the performance of the
$\tau$-EO algorithm~\cite{eo_jam}, which has been verified in the
problems studied so far~\cite{BoPe2,Dall,eo_prl} as exemplified in
Fig.~\ref{glasstau}.

\subsection{Theory of the EO Algorithm}
\label{EOtheory}
Stochastic local search heuristics are notoriously hard to
analyze. Some powerful results have been derived for the convergence
properties of SA in dependence of its temperature
schedule~\cite{Geman,Aarts2}, based on the well-developed knowledge of
equilibrium statistical physics (``detailed balance'') and Markov
processes. But predictions for particular optimization problems are
few and far between. Often, SA and GA, for instance, are analyzed on
simplified models (see Refs.~\cite{LundyMees86,Sorkin,C+F} for SA and
Ref.~\cite{Wegener} for GA) to gain insight into the workings of a
general-purpose heuristic. We have studied EO on an appropriately designed model
problem and were able to reproduce many of the properties observed for
our realistic $\tau$-EO implementations. In particular, we found 
analytical results for the average convergence as a function
of $\tau$~\cite{eo_jam}.

In Ref.~\cite{eo_jam} we have considered a model consisting of $n$
a-priori independent variables. Each variable $i$ can take on only one
of, say, three fitness states, $\lambda_i=0$, -1, an -2, respectively
assigned to fractions $\rho_0$, $\rho_1$, and $\rho_2$ of the
variables, with the optimal state being $\lambda_i=0$ for all $1\leq
i\leq n$, i.~e. $\rho_0=1$, $\rho_{1,2}=0$ and cost
$C=-\sum_i\lambda_i/n=\sum_{\alpha=0}^2\alpha\rho_{\alpha}=0$,
according to Eq.~(\ref{costeq}). With this system, we can model the
dynamics of local search for hard problems by ``designing'' an
interesting set of flow equations for ${\bf\rho}(t)$ which can mimic
a complex search space through energetic or entropic barriers, for
instance~\cite{eo_jam}. These flow equations specify what fraction of
variables transfer from on fitness state to another given that a
variable in a certain state is updated. The update probabilities are
easily derived for $\tau$-EO, giving a highly nonlinear dynamic
system. Other local searchs may be studied in this model for
comparison~\cite{Frank}.

\begin{figure}
\vskip 1.9in \includegraphics{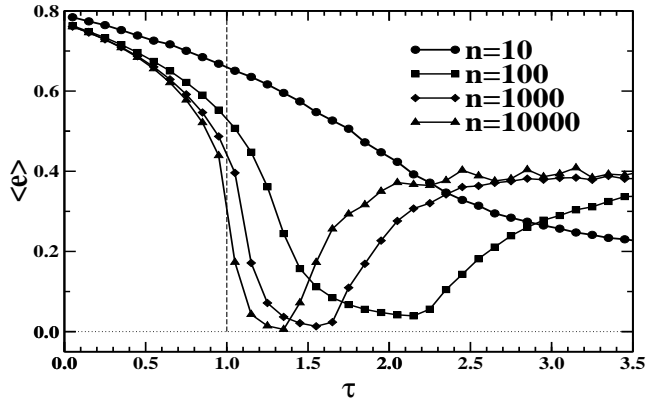}
\caption{Plot of the cost $\left< C\right>$ averaged over many
$\tau$-EO runs as a function of $\tau$ for $n=10$, 100, 1000, and
10000 from Ref.~\protect\cite{eo_jam}. It reaches a minimum with
$\left<C\right>\approx0$ at a value
near the prediction for $\tau_{\rm opt}\approx3.5$, 2.1, 1.6, and 1.4
[from Eq.~(\ref{tauopteq}) with $A\approx4$ and higher-order
corrections], and rises sharply beyond that, similar to empirical
findings, see Figs.~\protect\ref{glasstau}a-b. }
\label{jamtau}
\end{figure}

A particular design that allows the study of $\tau$-EO for a generic
feature of local search is suggested by the close analogy between
optimization problems and the low-temperature properties of spin
glasses~\cite{MPV}: After many update steps most variables freeze into
a near-perfect local arrangement and resist further change, while a
finite fraction remains frustrated in a poor local
arrangement~\cite{PSAA}. More and more of the frozen (slow) variables
have to be dislocated collectively to accommodate the frustrated
(fast) variables before the system as a whole can improve its
state. In this highly correlated state, slow variables block the
progression of fast variables, and a ``jam'' emerges. And our
asymptotic analysis of the flow equations for a jammed system indeed
reproduces key features previously conjectured for EO from the
numerical data for real optimization problems. Especially, it predicts
for the value $\tau$ at which the cost is minimal for a given runtime,
\begin{eqnarray}
\tau_{\rm opt}\sim1+{A\over\ln n}\quad (n\to\infty),
\label{tauopteq}
\end{eqnarray}
where $A>0$ is some implementation specific constant. This result was
found empirically before in Refs.~\cite{BoPe2,eo_prl}. The behavior of
the average cost $\left<C\right>$ as a function of $\tau$ for this model is shown
in Fig.~\ref{jamtau}, which verifies Eq.~(\ref{tauopteq}).

This model provides the ideal setting to probe deeper into the
properties of EO, and to compare it with other local search
methods. Similarly, EO can be analyzed in terms of a homogeneous
Markov chain~\cite{Feller,Jerrum}, although little effort has been
made in this direction yet (except for Ref.~\cite{VJ1}). Such
theoretical investigations go hand-in-hand with the experimental
studies to provide a clearer picture of the capabilities of EO.

\subsection{Comparison with other Heuristics}
\label{comparison}
As part of this project, we will often compare or combine EO with
Meta-heuristics \cite{Osman} and problem specific methods
\cite{Ausiello}. (This is also an important part of the educational
purpose of this proposal.) As we will show, EO provides an alternative
philosophy to the canon of heuristics. But these distinctions do not
imply that any of the methods are fundamentally better or worse. To
the contrary, their differences improve the chances that at least one
of the heuristics will provide good results on some particular problem
when all others fail! At times, best results are obtained by {\em
hybrid\/} heuristics \cite{Rayward,Voss,Reeves}.  The most apparent
distinction between EO and other methods is the need to define local
cost contributions for each variable, instead of only a global
cost. EO's capability seems to derive from its ability to access
this local information directly.

\highlite{Simulated Annealing (SA):} SA \cite{Science} emulates the
behavior of frustrated systems in {\em thermal equilibrium:} if one
couples such a system to a heat bath of adjustable temperature, by
cooling the system slowly one may come close to attaining a state of
minimal energy (i.~e. cost).  SA accepts or rejects local changes to a
configuration according to the  Metropolis algorithm \cite{MRRTT} at a
given temperature, enforcing equilibrium dynamics (``detailed
balance'') and requiring a carefully tuned ``temperature schedule''
\cite{Aarts1,Aarts2}

In contrast, EO drives the system {\em far from equilibrium:} aside
from ranking, it applies no decision criteria, and new configurations
are accepted indiscriminately. Instead of tuning a schedule of
parameters, EO often requires few choices.  It may appear that EO's
results should resemble an ineffective random search, similar to SA at
a fixed but finite temperature \cite{D+S,Fielding}. But in fact, by
persistent selection against the worst fitnesses, EO quickly
approaches near-optimal solutions. Yet, large fluctuations remain at
late runtimes (unlike in SA, see Fig.~\ref{runtime} or
Ref.~\cite{JohnsonGBP}) to escape deep local minima and to access new
regions in configuration space.

In some versions of SA, low acceptance rates near freezing are
circumvented using a scheme of picking trials from a rank-ordered list
of possible moves \cite{Greene} (see Chap.~2.3.4 in
Ref.~\cite{Reeves}), derived from continuous-time Monte Carlo methods
\cite{BKL}. Like in EO, every move gets accepted. But these moves are
based on an outcome-oriented ranking, favoring downhill moves but
permitting (Boltzmann-)limited uphill moves. On the other hand, in EO
the ranking of variables is based on the current, not the future,
state of each variable, allowing for unlimited uphill moves.

\highlite{Genetic Algorithms (GA):} Although similarly motivated by
evolution (with deceptively similar terminology, such as ``fitness''),
GA \cite{Holland,Goldberg} and EO algorithms have hardly anything in
common. GAs, mimicking evolution on the genotypical level, keep track
of entire ``gene pools'' of configurations and use many tunable
parameters to select and ``breed'' an improved generation of
solutions.  By comparison, EO, based on competition at the
phenomenological level of ``species,'' operates only with local
updates on a single configuration, with improvements achieved by
persistent elimination of bad variables. EO, SA, and other
general-purpose heuristics use a local search. In contrast, in GA
cross-over operators perform global exchanges on
a pair of configurations.

\highlite{Tabu-Search (TS):} TS performs a memory-driven local search
procedure that allows for limited uphill moves based on scoring recent
moves \cite{Glover,Reeves,Aarts3}. Its memory permits escapes from
local minima and avoids recently explored configurations. It is
similar to EO in that it may not converge ($S_{\rm best}$ has to be
kept!), and that moves are ranked. But the uphill moves in TS are
limited by tuned parameters that evaluate the memory. And, as for SA
above, rankings and scoring of moves in TS are done on the basis of
anticipated outcome, not on current ``fitness'' of individual
variables.

\section{EO-Implementations and Results}
\label{EOapplications}
We have conducted a whole series of projects to demonstrate the
capabilities of simple implementations in obtaining near-optimal
solutions for the GBP \cite{BoPe1,BoPe2,EOperc}, the 3-coloring of
graphs \cite{eo_prl,BGIP}, and the Ising spin-glass problem
\cite{eo_prl} (a model of disordered magnets that maps to a MAX-CUT
problem \cite{DIMACS7}). In each case we have studied a statistically
relevant number of instances from an ensemble with up to $10^4$
variables, chosen from ``Where the really hard problems
are'' \cite{AI}.  These results are discussed in the following.

\subsection{Graph Bipartitioning}
In Table~\ref{tab1} we summarize early results of our $\tau$-EO
implementation for the GBP on a testbed of graphs with $n$ as large as
$10^5$. Here, we use $\tau=1.4$ and the best-of-10 runs.  On each
graph, we used as many update steps $t$ as appeared productive for EO
to reliably obtain stable results.  This varied with the
particularities of each graph, from $t=2n$ to $200n$, and the reported
runtimes are influenced by this.
\begin{table}
\caption{Best cutsizes (and allowed runtime) for a testbed of large
graphs.  GA results are the best reported~\protect\cite{MF1} (at
300MHz). $\tau$-EO results are from our runs (at 200MHz), out-pacing the GA results by almost an order of magnitude for large $n$ .  Comparison
data for three of the large graphs are due to results from spectral heuristics
in Ref.~\protect\cite{HL} (at 50MHz). METIS is a partitioning program
based on hierarchical reduction instead of local
search~\protect\cite{METIS}, obtaining extremely fast deterministic
results (at 200MHz).  }
\begin{tabular}{l@{}r|r@{}lr@{}lr@{}lr@{}l}
\hline\hline \multicolumn{2}{l}{Large Graph\hfil $n$} & \multicolumn{2}{c}{GA}
& \multicolumn{2}{c}{$\tau$-EO}
&\multicolumn{2}{c}{Ref.~\protect\cite{HL}} &
\multicolumn{2}{c}{p-METIS} \\ \hline {\em Hammond\/}& 4720 & 90 & (1s) & 90 & (42s) & 97 & (8s) & 92 & (0s) \\ {\em
Barth5\/} & 15606 & 139 & (44s) & 139 & (64s) & 146 &
(28s) & 151 & (0.5s)\\ {\em Brack2\/} & 62632 & 731 &
(255s) & 731 & (12s) & \multicolumn{2}{c}{---} & 758 & (4s) \\ {\em
Ocean\/} & 143437 & 464 & (1200s) & 464 & (200s) & 499
& (38s) & 478 & (6s)\\ \hline\hline
\end{tabular}
\label{tab1}
\end{table}

\begin{figure}
\vskip 2.0truein \includegraphics{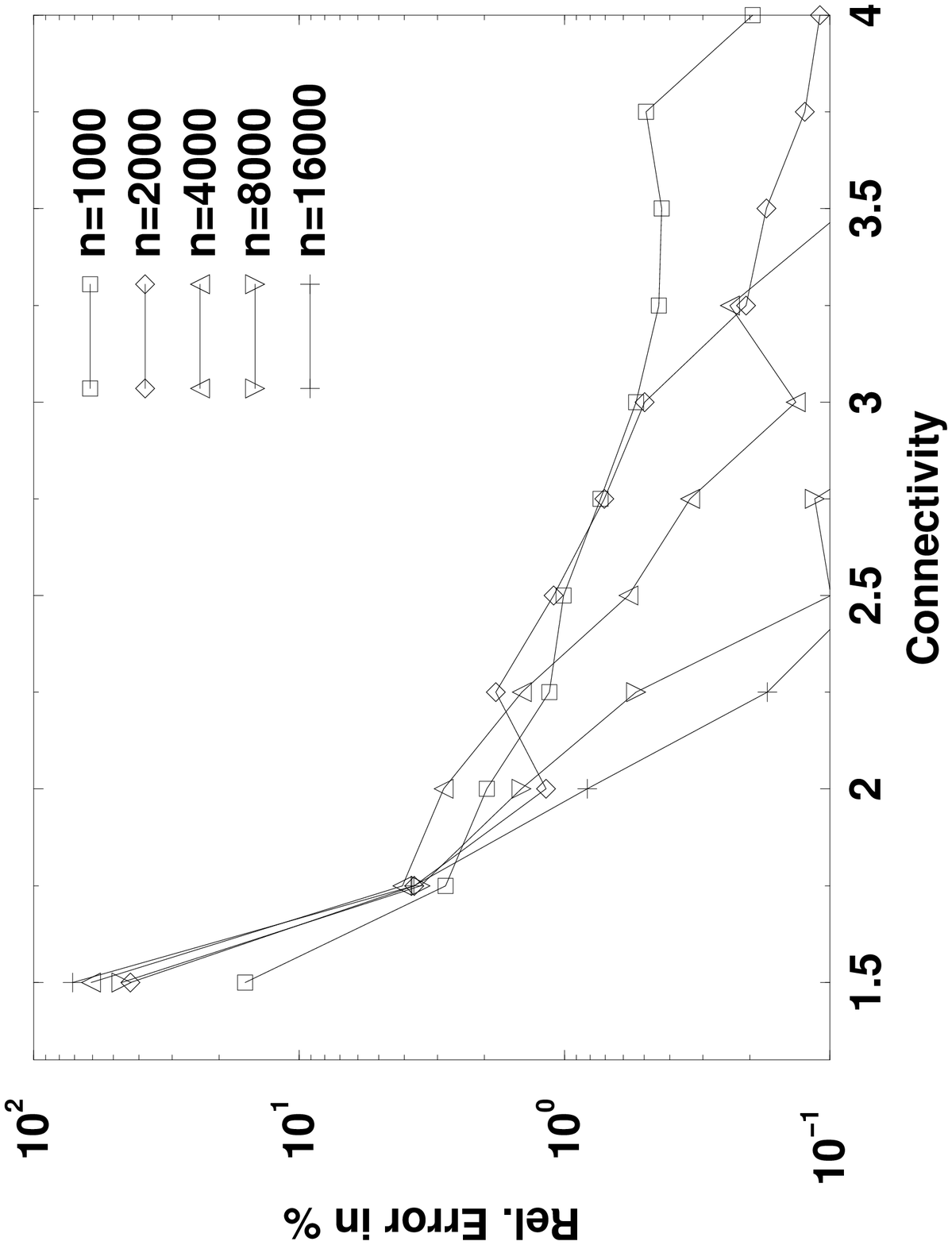} \includegraphics{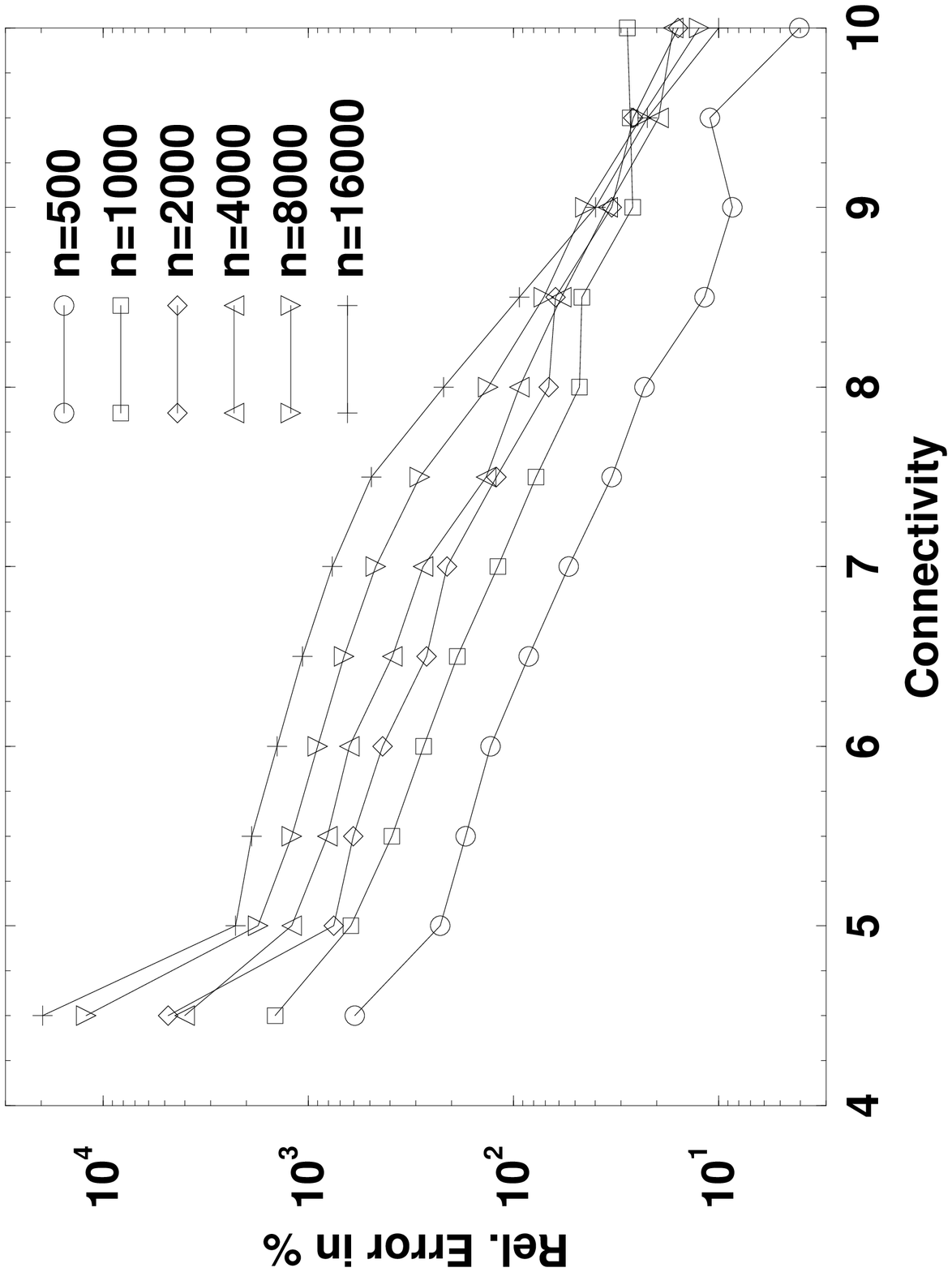}
\caption{Plot of the error in the best result of SA relative to EO's
on identical instances of  random graphs (left) and geometric graphs (right)
as function of the average connectivity $c$.  The critical points for
the GBP are at $c=2\ln2=1.386$ for random graphs and
at $c\approx 4.5$ for geometric graphs.  SA's error relative to EO near the
critical point in both cases rises with $n$.}
\label{perco}
\end{figure}

In an extensive numerical study on random and geometric graphs
\cite{EOperc} we have shown that $\tau$-EO outperforms SA
significantly near phase transitions, where
cutsizes first become non-zero.  To this end, we have compared the
averaged best results obtained for both methods for a large number of
instances for increasing $n$ at a fixed parameter setting. For EO, we have used the algorithm for GBP described in Sec.~\ref{taueoalgo} at $\tau=1.4$. For SA, we have used the algorithm developed  by Johnson \cite{JohnsonGBP} for GBP, with a geometric temperature schedule and a temperature length of $64n$ to equalize runtimes between EO and SA. Both programs used the same data structure, with EO requiring a small extra overhead for sorting the fitness of variables in a heap \cite{BoPe1}. Clearly, since each update leads to a move and entails some sorting, individual EO updates take much longer than an SA trial step. Yet, as Fig.~\ref{perco} shows, SA gets rapidly worse near
the phase transition relative to EO, at equalized CPU-time.

Studies on the average rate of convergence toward better-cost
configurations as a function of runtime $t$ indicate power-law
convergence, roughly like $\left<C(S_{\rm best})\right>_t\sim \left<C(S_{\rm
min})\right>+A\,t^{-0.4}$ \cite{BoPe2}, also found by Ref.~\cite{Dall}. Of
course, it is not easy to assert for graphs of large $n$ that those
runs in fact converge closely to the optimum $C(S_{\rm min})$, but
finite-size scaling analysis for random graphs justifies that
expectation \cite{BoPe2}.

\subsection{Graph Coloring}
 An instance in graph coloring consists of a graph with $n$ vertices,
some of which are connected by edges, just like in the GBP. We have
considered the problem of MAX-$K$-COL: given $K$ different colors to
label the vertices, find a coloring of the graph that minimizes the
number of ``monochromatic'' edges that connect vertices of identical
color.

For MAX-$K$-COL we define the fitness as $\lambda_i=-b_i/2$, like for
the GBP, where $b_i$ is the number of monochromatic edges emanating
from vertex $i$. Since there are no global constraints, a simple
random reassignment of a new color to the selected variable $x_j$ is a
sufficient local-search neighborhood.

We have studied the MAX-3-COL problem near its phase transition, where
the hardest instances reside~\cite{HH,Cheeseman,Culberson,AI}. In
Ref.~\cite{Cheeseman} the phenomena of phase transition has been
studied first for $3$- and $4$-COL. Here, we used EO to completely
enumerate {\em all\/} optimal solutions $S_{\rm min}$ near the
critical point for $3$-COL of random graphs. Instances of random
graphs typically have a high ground-state degeneracy, i.~e. possess a
large number of equally optimal solutions $S_{\rm min}$.  In
Ref.~\cite{Monasson} it was shown that at the phase transition of
$3$-SAT the fraction of {\em constrained variables,} i.~e. those that
are found in an identical state in almost {\em all\/} $S_{\rm min}$,
discontinuously jumps to a non-zero value. It was conjectured that
this {\em first-order\/} phase transition in this ``backbone'' is a
general phenomenon for NP-hard optimization problems.

\begin{figure}
\vskip 1.8truein \includegraphics{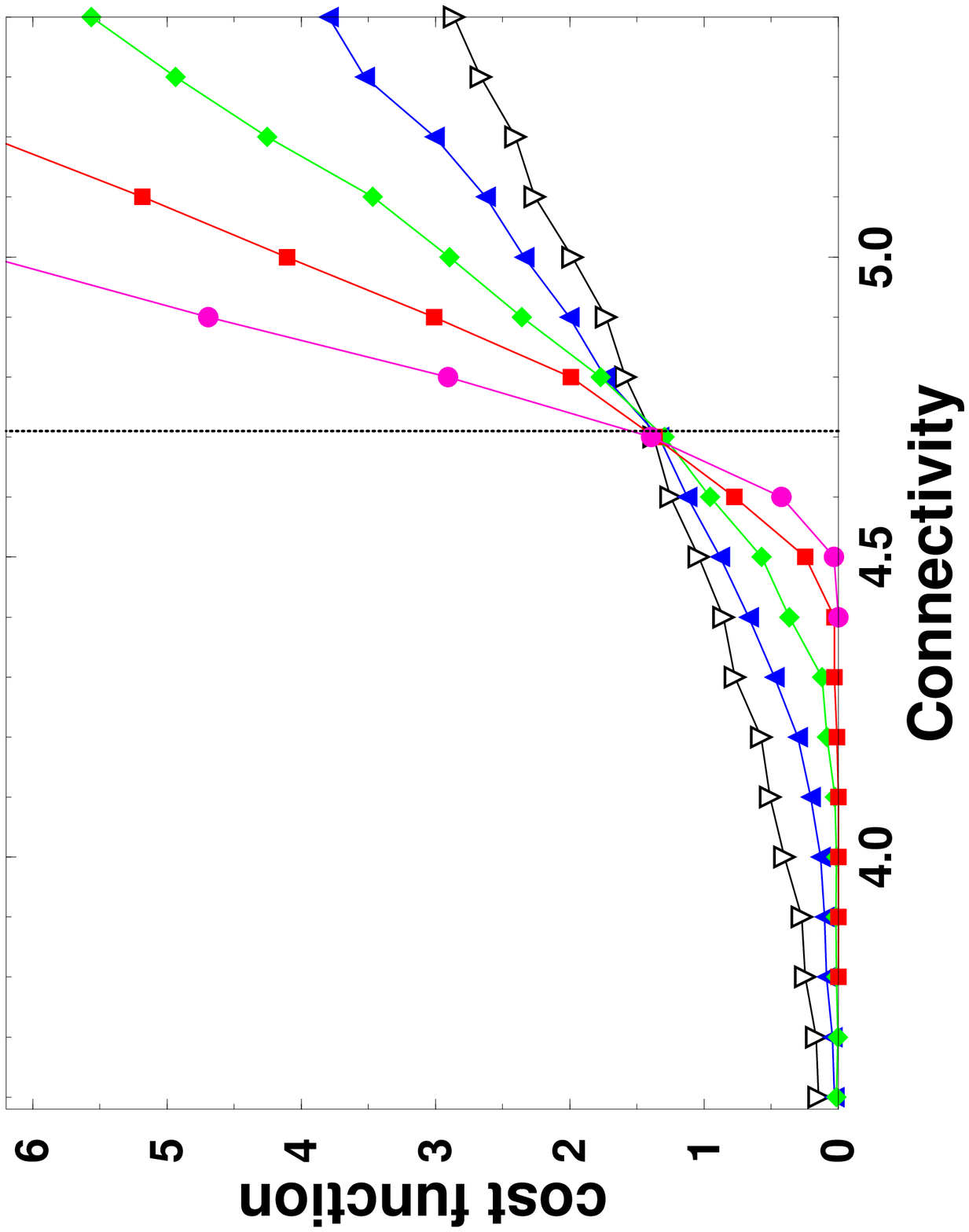} \includegraphics{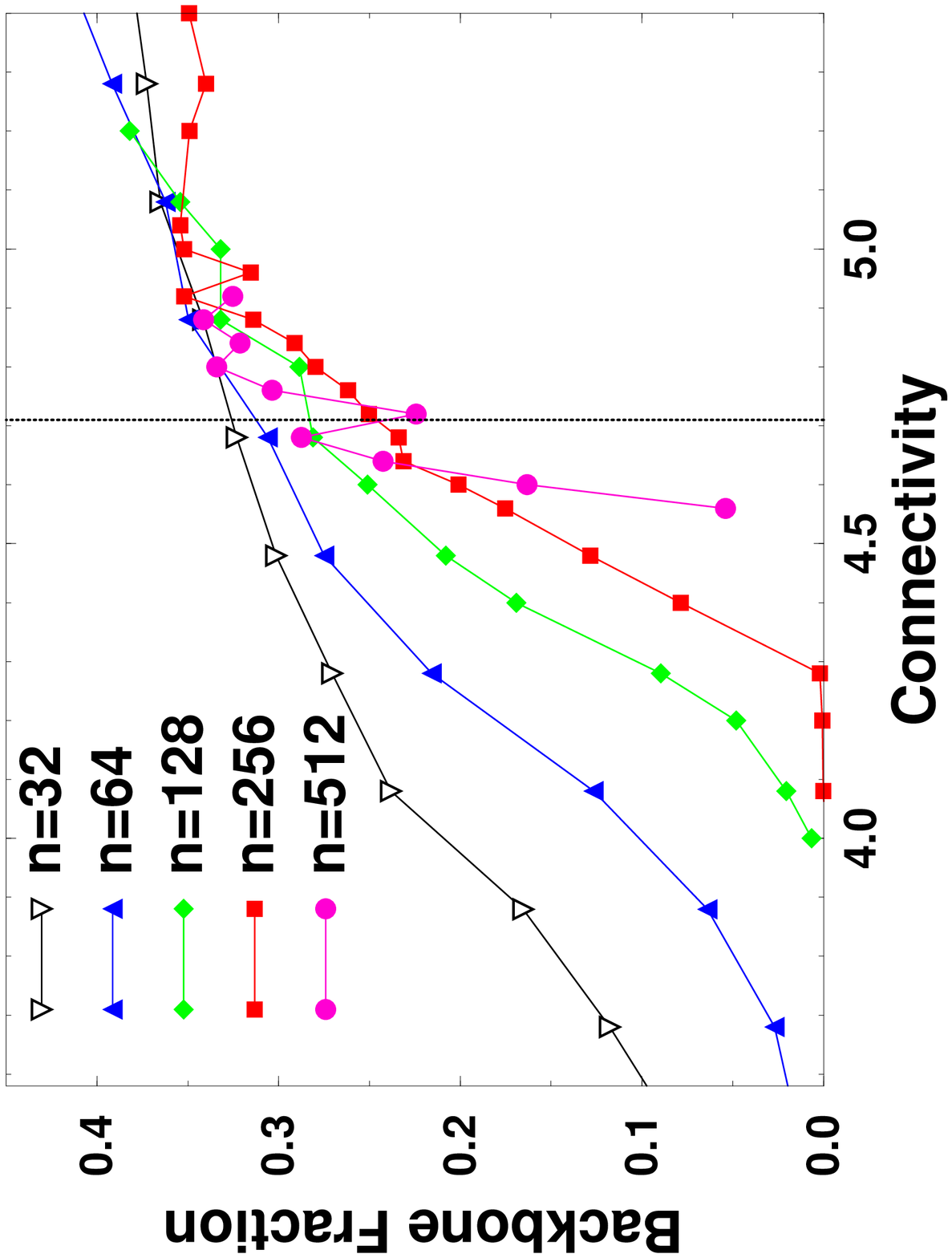}
\includegraphics{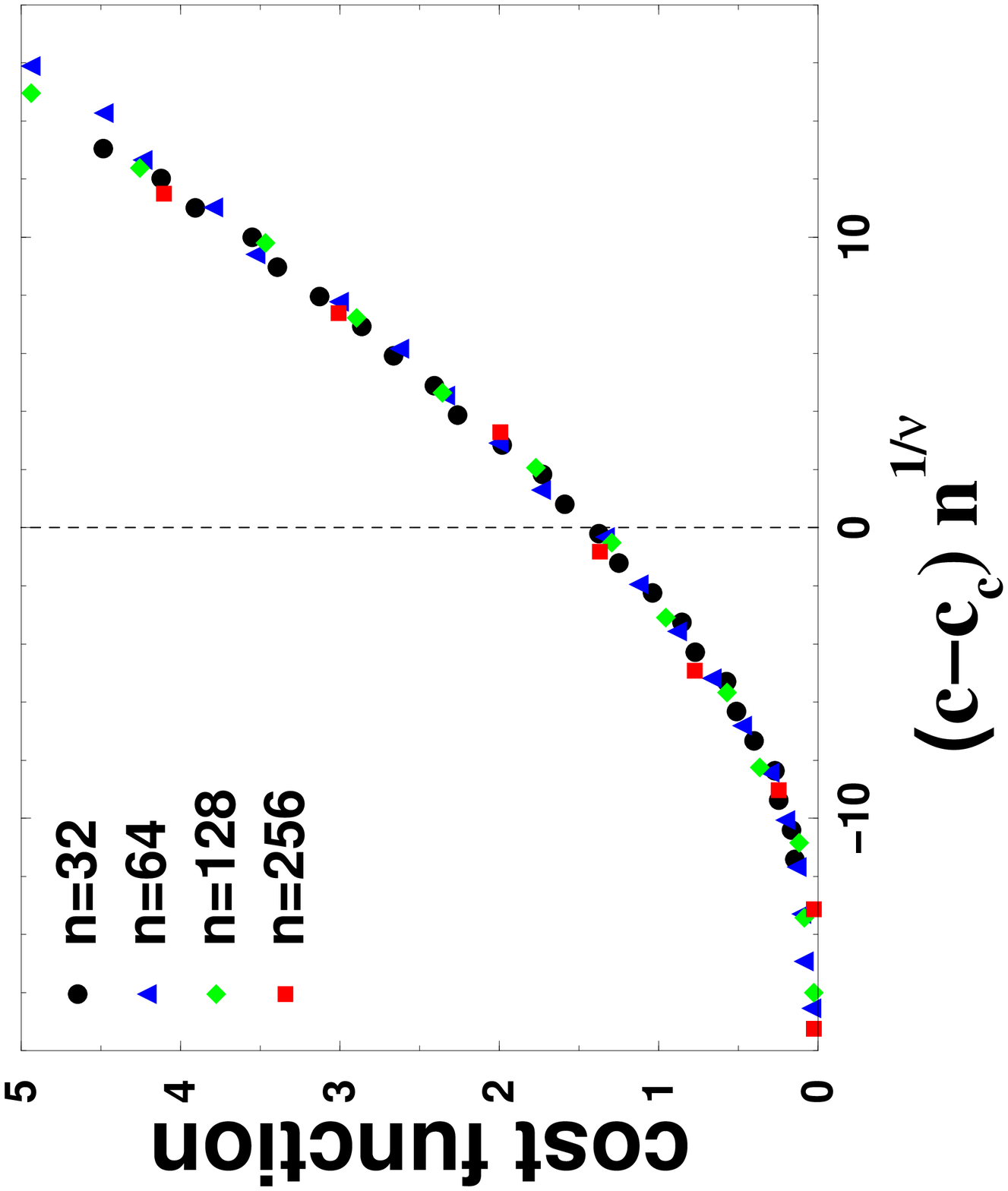}
\caption{Plot of the average cost (left) and  of the backbone fraction (right)
as a function of the average connectivity $c$ for random graph
3-coloring. The data collapse according to Eq.~(\protect\ref{scalingeq}) in the insert on the left predicts a critical
point for random graphs at $c_{\rm crit}\approx4.72$
(indicated by a vertical line) and $\nu=1.53(5)$. We generated at each
value of $c$ $10000$, $5000$, $1300$, $650$, and $150$ instances for
$n=32$, 64, 128, 256, and 512, respectively. }
\label{3colplot}
\end{figure}

 To test the conjecture for the 3-COL, we generated a large number of
random graphs and explored $\Omega$ for as many ground states as EO
could find. (We fixed runtimes well above the times needed to saturate
the set of all $S_{\rm min}$ in repeated trials on a testbed of
exactly known instances.) For each instance, we measured the optimal
cost and the backbone fraction of fixed pairs of vertices. The results
in Fig.~\ref{3colplot} allow us to estimate precisely the location of
the transition and the scaling behavior of the cost function.
 With a  finite-size
scaling ansatz to ``collapse'' the data for the average ground-state
cost onto a single scaling curve,
\begin{eqnarray}
\left< C\right>\sim n f\left[\left(c-c_{\rm crit}\right)n^{1/\nu}\right],
\label{scalingeq}
\end{eqnarray}
it is possible to extract precise estimates for the location of the
transition $c_{\rm crit}$ and the scaling window exponent~$\nu$.

\subsection{``Spin Glasses'' (or MAX-CUT)} Of significant physical
relevance are the low temperature properties of ``spin glasses''
\cite{MPV}, which are closely related to MAX-CUT
problems~\cite{DIMACS7}. EO was originally designed with applications
to spin glasses in mind, and some of its most successful results were
obtained for such systems~\cite{eo_prl}.
Many physical and classic combinatorial optimization problems
(Matching, Partitioning, Satisfiability, or the Integer Programming
problem below) can be cast in terms of a spin glass~\cite{MPV}.

A spin glass consists of a lattice or a graph with a spin variable
$x_i\in\{-1,1\}$ placed on each vertex $i$, $1\leq i\leq n$. Every
spin is connected to each of its nearest neighbors $j$ via a fixed
bond variable $J_{i,j}$, drawn at random from a distribution of zero
mean and unit variance. Spins may be coupled to an arbitrary external
field $h_i$.  The optimization problem consists of finding minimum
cost states $S_{\rm min}$ of the ``Hamiltonian''
\begin{eqnarray}
C(S)=H(x_1,\ldots,x_n)=-{1\over2}\sum_{i,j}J_{i,j}x_ix_j-\sum_ix_i
h_i.
\label{spineq1}
\end{eqnarray}
Arranging spins into an optimal configuration is hard due to
``frustration:'' variables that will, individually or collectively, never be able to satisfy all constraints imposed on them. The cost function in
Eq.~(\ref{spineq1}) is equivalent to integer
quadratic programming problems \cite{DIMACS7}.

We simply define as fitness the local cost contribution for each spin,
\begin{eqnarray}
\lambda_i=x_i\left({1\over2}\sum_j J_{i,j}x_j+h_i\right),
\end{eqnarray}
and Eq.~(\ref{spineq1}) turns into Eq.~(\ref{costeq}). A single spin
flip provides a sufficient neighborhood for this problem.  This formulation
trivially extends to higher than quadratic couplings.

We have run this EO implementation for a spin glass with $h_i\equiv0$ and
 random $J_{i,j}=\pm1$ for nearest-neighbor bonds on a cubic lattice
 \cite{eo_prl}. We used $\tau=1.15$ on a large number of realizations
 of the $J_{ij}$, for $n=L^3$ with $L=5,6,7,8,9,10,12$. For
 each instance, we have run EO with 5 restarts from random initial
 conditions, retaining only the lowest energy state obtained, and then
 averaging over instances. Inspection of the results for convergence
 of the genetic algorithms in Refs.~\cite{Pal1,Houdayer} suggest a
 computational cost per run of at least $O(n^3--n^4)$ for consistent
 performance. Indeed, using $\sim n^4/100$ updates enables EO to
 reproduce its lowest energy states on about 80\% to 95\% of the
 restarts, for each $n$. Our results are listed in
 Tables~\ref{table1}. A fit of our data for the energy per spin,
 $e(n)=<C>_n/n$, $C(s)$ defined in Eq.~(\ref{spineq1}), with $e(n)=e(\infty)+{\rm const}/n$ for $n\to\infty$ predicts
 $e(\infty)=1.7865(3)$, consistent with the findings of
 Refs.~\cite{Pal1,Hartmann_d3}, providing independent confirmation of
 those results with far less parameter tuning.

\begin{table}
\caption{ EO approximations to the average ground-state energy per
spin $e(n)$ of the $\pm J$ spin glass in $d=3$, compared with GA
results from Refs.~\protect\cite{Pal1,Hartmann_d3}. For each size
$n=L^3$ we have studied a large number $I$ of instances. Also shown is
the average time $t$ (in seconds) needed for EO to find the presumed
ground state on a 450MHz Pentium. (As for a normal distribution, for increasing $n$ fewer instances are needed to obtain similar error bars.)}
\begin{tabular}{rrlcll}
\hline\hline
$L$ & $I$ & $e(n)$ & $t$ & Ref.~\protect\cite{Pal1} &
Ref.~\protect\cite{Hartmann_d3} \\
\hline
     3  &40100  & -1.6712(6) &0.0006  &-1.67171(9) & -1.6731(19)\\
     4  &40100  & -1.7377(3) &0.0071  &-1.73749(8) &-1.7370(9)\\
    5  &28354  & -1.7609(2) &0.0653  &-1.76090(12)&-1.7603(8) \\
    6  &12937  & -1.7712(2) &0.524  &-1.77130(12)& -1.7723(7)\\
    7  & 5936  & -1.7764(3) &3.87  & -1.77706(17) &  \\
    8  & 1380  & -1.7796(5) &22.1  &-1.77991(22)&-1.7802(5)\\
    9  &  837  & -1.7822(5) &100. &&\\
   10  &  777  & -1.7832(5) &424.  &-1.78339(27)  &-1.7840(4)\\
   12  &   30  & -1.7857(16) &9720.  &-1.78407(121) &-1.7851(4)\\
\hline\hline
\end{tabular}
\label{table1}
\end{table}

To gauge EO's performance for larger $n$, we have run our
implementation also on two $3d$ lattice instances, $toruspm$3-8-50 and
$toruspm$3-15-50, with $n=8^3$ and $n=15^3$, considered in the 7th
DIMACS challenge for semi-definite problems \cite{DIMACS7}. Bounds
\cite{Juenger} on the ground-state cost established for the larger
instance are $C_{\rm lower}=-6138.02$ (from semi-definite programming)
and $C_{\rm upper}=-5831$ (from branch-and-cut). EO found $C(S_{\rm
best})=-6049$ (or $e=C/n=-1.7923$), a significant improvement on the
upper bound and already lower than $e(\infty)$ from above. Furthermore,
we collected $10^5$ such states, which roughly segregate into 3
clusters with a mutual Hamming distance of at least 100 distinct
spins.  For the smaller instance the bounds given are -922 and -912,
resp., while EO finds -916 (or $C/n=-1.7891$). While this run
(including sampling degenerate states!) took only a few minutes of CPU
(at 800MHz), the results for the larger instance require about 16
hours.

\begin{acknowledgments}
The authors wish to acknowledge financial support from the Emory University Research Committee and from a Los Alamos LDRD grant.
\end{acknowledgments}


\end{document}